\documentclass[letterpaper,twocolumn,10pt]{article}
\usepackage{usenix2019_v3}

\usepackage{tikz}
\usepackage{amsmath,amssymb,amsfonts}
\usepackage{graphicx}
\usepackage{textcomp}
\usepackage{xcolor}
\usepackage{fancyhdr}
\usepackage{threeparttable} 
\usepackage{multirow}
\usepackage{graphicx}
\usepackage{tcolorbox}
\usepackage{enumitem}
\usepackage{booktabs}
\usepackage{subcaption}
\usepackage{pbox}
\usepackage{hyperref}
\usepackage{inconsolata}  

\usepackage{pifont}
\usepackage[linesnumbered,ruled,vlined]{algorithm2e}

\usepackage{listings}
\definecolor{problemblue}{RGB}{100,134,158}
\definecolor{idiomsgreen}{RGB}{0,162,0}
\definecolor{exercisebgblue}{rgb}{0,  .69,  .941}
\definecolor{deepgreen}{rgb}{0.0, 0.5, 0.0}
\definecolor{codegreen}{rgb}{0,0.6,0}
\definecolor{codegray}{rgb}{0.5,0.5,0.5}
\definecolor{codepurple}{rgb}{0.58,0,0.82}
\definecolor{backcolour}{rgb}{0.95,0.95,0.92}
\definecolor{redColor}{RGB}{255,0,0}
\definecolor{Gray}{gray}{0.1}

\definecolor{javapurple}{rgb}{0.5,0,0.35}
\definecolor{javagreen}{rgb}{0,0.4,0}

\lstdefinelanguage{customc}
{
  morekeywords={for, int, return},
  morecomment=[l]{//},
  morecomment=[s]{/*}{*/},
  morestring=[b]",
  basicstyle=\small\ttfamily,
  numbers=left, firstnumber=1, numberstyle=\tiny\color{gray},
  showstringspaces=false,
  escapeinside={(*@}{@*)},
  keywordstyle=\color{javapurple},
  columns=fullflexible,
  showlines=true,
  xleftmargin=0.5cm,
  moredelim=**[is][\color{NavyBlue}]{!}{!},
  moredelim=**[is][\color{OliveGreen}]{<_}{_>}
}
\lstdefinelanguage{python}{
  morekeywords={if, else, def, return, Synthesized, Invariant},
  morecomment=[l]{//},
  morecomment=[s]{/*}{*/},
  morestring=[b]",
  basicstyle=\small\ttfamily,
  numbers=left, firstnumber=1, numberstyle=\tiny\color{gray},
  showstringspaces=false,
  escapeinside={(*@}{@*)},
  keywordstyle=\color{javapurple},
  columns=fullflexible,
  showlines=true,
  xleftmargin=0.5cm,
  moredelim=**[is][\color{NavyBlue}]{!}{!},
  moredelim=**[is][\color{OliveGreen}]{<_}{_>}
}

\SetKwComment{Comment}{$\triangleright~$}{}

\usepackage{filecontents}
\newcommand{\todo}[1]{{\color{black}#1}}

\newcommand{\name}{\textit{QiMeng-Xpiler}}

\newcommand{\namewosmt}{\textit{QiMeng-Xpiler} w/o SMT}

\usepackage{authblk}
\usepackage{amssymb}  
\usepackage{amsfonts} 

\begin{document}

\date{}

\title{\Large \bf  QiMeng-Xpiler: Transcompiling Tensor Programs for \\Deep Learning Systems with a Neural-Symbolic Approach}


    
\author[1,2]{Shouyang Dong}
\author[3]{Yuanbo Wen}
\author[3]{Jun Bi}
\author[3]{Di Huang}
\author[3]{Jiaming Guo}
\author[1,2,3]{Jianxing Xu}
\author[1,2,3]{Ruibai Xu}
\author[3]{Xinkai Song}
\author[3]{Yifan Hao}
\author[4]{Ling Li}
\author[1]{Xuehai Zhou}
\author[2]{Tianshi Chen}
\author[3]{Qi Guo}
\author[3]{Yunji Chen\textsuperscript{\textsection}}
\affil[1]{University of Science and Technology of China}
\affil[2]{Cambricon Technologies}
\affil[3]{SKL of Processors,
Institute of Computing Technology,
Chinese Academy of Sciences}
\affil[4]{Institute of Software, Chinese Academy of Sciences}


\renewcommand\Authands{ and }

\maketitle

\footnotetext[1]{Yunji Chen is the corresponding author.}

\begin{abstract}
Heterogeneous deep learning systems (DLS) such as GPUs and ASICs have been widely deployed in industrial data centers, which requires to develop multiple low-level tensor programs for different platforms. An attractive solution to relieve the programming burden is to transcompile the legacy code of one platform to others. However, current transcompilation techniques struggle with either tremendous manual efforts or functional incorrectness, rendering “Write Once, Run Anywhere” of tensor programs an open question.

We propose a novel transcompiler, i.e., \textit{QiMeng-Xpiler}, for automatically translating tensor programs across DLS via both large language models (LLMs) and symbolic program synthesis, i.e., neural-symbolic synthesis. 
The key insight is \emph{leveraging the powerful code generation ability of LLM to make costly search-based symbolic synthesis computationally tractable}. Concretely, we propose multiple LLM-assisted compilation passes via pre-defined meta-prompts for program transformation. During each program transformation, efficient symbolic program synthesis is employed to repair incorrect code snippets with a limited scale. To attain high performance, we propose a hierarchical auto-tuning approach to systematically explore both the parameters and sequences of transformation passes.
Experiments on $4$ DLS with distinct programming interfaces, i.e., Intel DL Boost with VNNI, NVIDIA GPU with CUDA, AMD MI with HIP, and Cambricon MLU with BANG, demonstrate that \textit{QiMeng-Xpiler} correctly translates different tensor programs at the accuracy of 95\% on average, and the performance of translated programs achieves up to $2.0\times$ over vendor-provided manually-optimized libraries. As a result, the programming productivity of DLS is improved by up to $96.0\times$ via transcompiling legacy tensor programs.

\end{abstract}

\vspace{-10pt}
\section{Introduction}

Due to the ever-increasing demand for computation from neural network workloads, various deep learning systems (DLS), e.g., NVIDIA GPU with Tensor Core~\cite{NVTensorCore}, Google TPU~\cite{Jouppi17ISCA}, GraphCore IPU~\cite{graphcoreipu}, and Cambricon MLU~\cite{MLU}, have been deployed in data centers of cloud and internet service companies such as Microsoft~\cite{Microsoftcloud}, Google~\cite{Googlecloud}, and Amazon~\cite{Amazoncloud}. To fully exploit various DLS, it is required to develop multiple high-performance tensor programs, i.e., low-level implementations of tensor operators in deep learning, for different platforms, which are notoriously challenging because of the complicated architecture and programming models. Ideally, if we only need to write one copy of a program, and can run it on different platforms, that is, “Write Once, Run Anywhere”, it is feasible to address the programming problem in data centers with heterogeneous DLS.

\begin{figure}[t]
    \vspace{-0pt}
    \centering
    \includegraphics[width=1.0\linewidth]{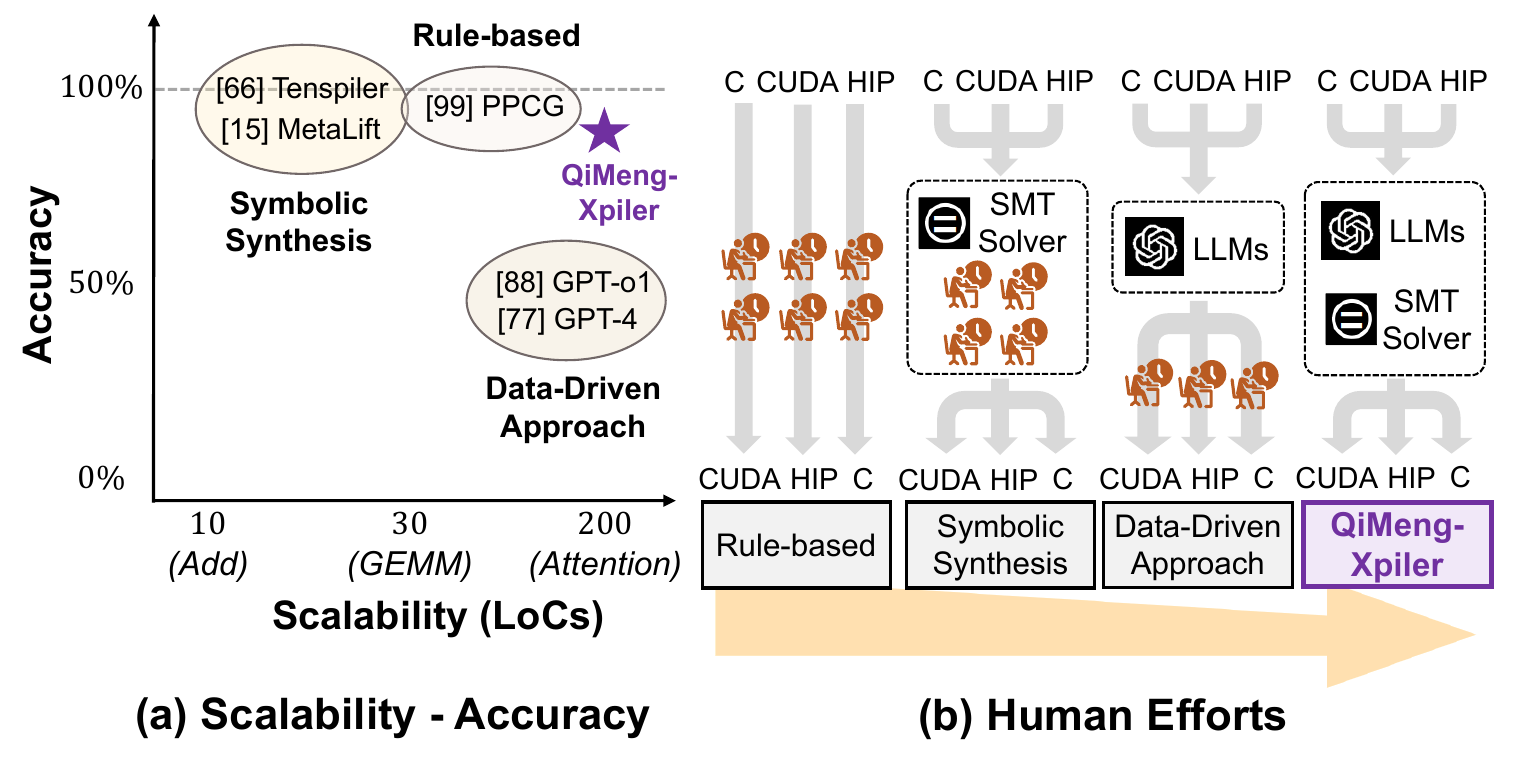}
    \caption{\small Comparing \name{} to existing transcompilation techniques in terms of (a) scalability-accuracy axis and (b) human efforts}
    \label{fig:intro_comparison}
\end{figure}

One of the most attractive solutions to achieve “Write Once, Run Anywhere” on heterogeneous platforms is source-to-source compilation, i.e., transcompilation, which automatically compiles the code written with one high-level language to another. Existing transcompilation efforts fall into three categories, i.e., rule-based, symbolic synthesis, and data-driven approaches. The rule-based approaches require experts to manually define a set of transformation rules, typically applied to Abstract Syntax Tree (AST), between different programming languages, and then employ pattern matching to parse the input source programs~\cite{johnson2022martini,10.1145/3554736}. For example, a CUDA-to-FPGA translator, i.e., FCUDA, defines a set of transformation rules for data communication, compute optimization, and parallelism mapping~\cite{PAPAKONSTANTINOU13TECS}. The symbolic synthesis approaches generate semantic-preserving target code from either domain-specific languages~\cite{Phitchaya14PLDI} or input/output examples~\cite{Jackson22PLDI}. Since they typically rely on a costly search-based SMT (Satisfiability Modulo Theories) solver~\cite{Z3Solver}, it is hard to scale to large-size general-purpose programs. The data-driven approaches, which typically train neural networks from a large amount of source code for code generation, have emerged recently~\cite{bahdanau2014neural,wu2016google,Shah18NIPS}. Notable examples include TransCoder~\cite{Roziere20NIPS},  StarCoder~\cite{li2023starcoder}, GPT-4~\cite{achiam2023gpt}, and OpenAI-o1~\cite{openaio1} which automatically translate programs written in different high-level languages such as C++, Java, and Python.

However, the above transcompilation techniques are not applicable to DLS due to their complicated architecture and programming models. Regarding the rule-based approaches, the huge architectural discrepancy between different platforms makes it infeasible to manually define efficient transformation rules. Regarding the symbolic synthesis approaches, in addition to their limited scalability, it is infeasible to handle different parallel semantics, e.g., SIMT in CUDA C and SIMD in BANG C~\cite{bangc}, for the SMT solver. Meanwhile, it also requires considerable manual efforts to accurately specify the input constraints for the SMT solver. 
Regarding the data-driven approaches, even with pre-trained large language models (LLMs), tensor program semantics cannot be fully preserved and thus the translation accuracy only achieves 29.6\%~\cite{li2022competition}, which requires considerable manual efforts for code correction.
In summary, existing transcompilation techniques for tensor programs face challenges such as tremendous manual efforts, limited scalability, or functional incorrectness, as shown in Figure~\ref{fig:intro_comparison}.

To automatically translate tensor programs across deep learning systems with a correctness guarantee, we propose to exploit both the \emph{flexibility of data-driven approaches}, specifically, via LLMs, and the \emph{soundness of symbolic synthesis approaches}. However, the challenge of collaboration between LLM and symbolic synthesis is three-fold: (1) it is difficult to generate highly accurate code via LLM due to the scarcity of tensor programs, (2) it is non-trivial to determine which code snippets should be formally generated by search-based symbolic synthesis, in order to balance the computational costs and achieved accuracy, and (3) performance optimizations are hard to specify either with LLM prompts or synthesis specifications, making the performance of generated code significantly lags behind that of human experts.

\subsection{Our Proposal}
To address the above challenges, we propose to build a novel transcompiler, i.e., \name{}, for translating tensor programs across DLS via both LLM (i.e., GPT-4~\cite{bubeck2023sparks}) and search-based symbolic synthesis. The key of \name{} is that \emph{the program translation is automatically conducted as a series of LLM-assisted transformation passes, where incorrect code snippets are repaired by small-scale symbolic synthesis, and the optimal transformation passes are identified via hierarchical auto-tuning}. The advantages of proposed neural-symbolic synthesis are: (1) the entire transcompilation is decomposed into multiple transformation passes via a chain of LLM prompting, instead of a single prompt in traditional LLM-assisted code generation, to significantly improve the accuracy, (2) the problem size for symbolic synthesis is constrained to a limited scale so that the SMT solver can handle it efficiently, and (3) the performance optimization is achieved by exploring both the parameters and sequences of transformation passes.

\begin{table*}[t]
\vspace{-10pt}
    \caption{Comparison of deep learning systems (Intel DL Boost, NVIDIA GPU, AMD MI, and Cambricon MLU) and their programming interfaces. The characteristics fall into three categories: Parallelism, Memory Hierarchy, and Specialized Intrinsics.}
    \label{tab:my_label}
    \centering
    \scriptsize
    \vspace{-5pt}
\begin{tabular}{ccll}
\toprule
\textbf{Platforms}             & \textbf{Interfaces}     & \multicolumn{1}{c}{\textbf{Categories}}         & \multicolumn{1}{c}{\textbf{Examples}}                                                                              \\ 
\midrule
Intel DL Boost               & \begin{tabular}[c]{@{}c@{}}C with\\ VNNI extensions\end{tabular}        & Specialized Intrinsic                & \texttt{\_mm\_dpbusds\_epi32(...), \_mm512\_dpbusd\_epi32(...)} \\                                                                        
\midrule
\multirow{4}{*}{\begin{tabular}[c]{@{}c@{}}NVIDIA GPU\\ with Tensor Core\end{tabular}}    & \multirow{4}{*}{CUDA C} & Parallelism                                   & \texttt{blockIdx, threadIdx}                                                                                       \\ \cmidrule{3-4} 
                               &                         & \multirow{2}{*}{Memory Hierarchy}             & \texttt{registers, \_\_shared\_\_, \_\_global\_\_}                                                                 \\ \cmidrule{4-4} 
                               &                         &                                               & \begin{tabular}[c]{@{}l@{}}\texttt{matrix\_a}, \texttt{matrix\_b}, and \texttt{accumulator} in Tensor Core fragments\end{tabular} \\ \cmidrule{3-4} 
                               &                         & Specialized Intrinsics                  & \texttt{wmma::mma\_sync(d, a, b, c)}                                                                               \\ \midrule
\multirow{4}{*}{\begin{tabular}[c]{@{}c@{}}AMD MI\\ with Matrix Core\end{tabular}}        & \multirow{4}{*}{HIP}    & Parallelism                                   & \texttt{blockIdx, threadIdx}                                                                                 \\ \cmidrule{3-4} 
                               &                         & \multirow{2}{*}{Memory Hierarchy}             & \texttt{registers, \_\_shared\_\_, \_\_global\_\_}                                                                 \\ \cmidrule{4-4} 
                               &                         &                                               & \begin{tabular}[c]{@{}l@{}}\texttt{matrix\_a}, \texttt{matrix\_b}, and \texttt{accumulator} in Matrix Core fragments\end{tabular} \\ \cmidrule{3-4} 
                               &                         & Specialized Intrinsics                  & \texttt{d = \_\_builtin\_amdgcn\_mfma\_f32\_16x16x4f32(a, b, c, ...)}                                              \\ \midrule
\multirow{6}{*}{Cambricon MLU} & \multirow{6}{*}{BANG C} & \multirow{2}{*}{Parallelism}                  & \texttt{taskId} for task-level parallelism                                                                                                    \\ \cmidrule{4-4} 
                               &                         &                                               & \texttt{clusterId, coreId}    for multi-core parallelism                                                                                     \\ \cmidrule{3-4} 
                               &                         & \multirow{2}{*}{Memory Hierarchy}             & \texttt{registers, \_\_mlu\_shared\_\_, \_\_mlu\_device\_\_}                                                       \\ \cmidrule{4-4} 
                               &                         &                                               & \texttt{\_\_nram\_\_, \_\_wram\_\_}                                                                                \\ \cmidrule{3-4} 
                               &                         & Specialized Intrinsics & \texttt{\_\_bang\_mlp(...)}, \texttt{\_\_bang\_conv(...)}                                                                                                                   \\ \bottomrule
\end{tabular}

\vspace{-10pt}   
\end{table*}

Concretely, by considering the high parallelism, memory hierarchy, and specialized ISA of DLS, the proposed approach consists of three categories of transformation, i.e., \emph{sequentialization/parallelization}, \emph{memory conversion}, and \emph{(de)tensorization}. Each transformation pass is first performed with LLM-based code transformation, then validated with unit-test, and finally repaired with an SMT-solver if necessary. 
Then, a hierarchical auto-tuning approach consisting of the \emph{intra-pass} and \emph{inter-pass} auto-tuning is conducted to improve the performance. The intra-pass auto-tuning uses brute-force search to find the optimal parameters (e.g., tiling sizes) for program transformation, and the inter-pass auto-tuning uses MCTS (Monte Carlo Tree Search)~\cite{6145622} to determine the optimal sequence of passes with maximized performance.

To demonstrate the generality of \name{}, we conduct experiments on 4 different DLS with their interfaces, i.e.,  Intel DL Boost with
VNNI intrinsics, NVIDIA GPU with CUDA C, AMD MI with HIP code, and
Cambricon MLU with BANG C. Experimental results show that \name{} correctly translates different programs at the accuracy of 95\% on average, 
and the performance of translated programs achieves up to $2.0\times$ over \todo{vendor-provided manually-optimized libraries including cuDNN/cuBLAS~\cite{NVCUDNN,NVCUBLAS} and oneDNN~\cite{oneDNN}}.
Moreover, the programming productivity of $2$ representative DLS, i.e., NVIDIA GPU and MLU, is improved by up to $34.3\times$ and $96.0\times$, respectively, via transcompiling legacy tensor programs.

\subsection{Key Contributions}
To our best knowledge, this work is \emph{the first to automatically translate tensor programs of deep learning systems with different programming models}. This paper makes the following contributions:
\begin{itemize}

\item {\textbf{Neural-symbolic program synthesis}}. We propose to use LLMs for generating high-level program sketches via pre-defined meta-prompts and repair the incorrect low-level details through SMT-based symbolic synthesis with limited scale, which achieves \emph{fully automatic program translation with a correctness guarantee}.

\item{\textbf{Hierarchical performance auto-tuning}}. We propose a hierarchical auto-tuning approach, where the \emph{intra-pass} and \emph{inter-pass} auto-tuning are used for exploring pass parameters and pass sequences, respectively, to maximize the performance of synthesized programs.

\item{\textbf{Extensive evaluation}}. We conduct extensive evaluations on 4 different deep learning systems, and experimental results on accuracy, execution performance, and productivity improvement well demonstrate the effectiveness and efficiency of this work.

\end{itemize}

\vspace{-10pt}
\section{Background and Motivation}
\vspace{-5pt}
\subsection{Background}
\subsubsection{Programming Deep Learning Systems}

Deep learning systems (DLS) have their own programming languages, providing adaptable interfaces for hardware manipulation, such as CUDA C for NVIDIA GPU, HIP for AMD MI, BANG C for Cambricon MLU, and VNNI instruction extensions for Intel DL Boost CPU.
However, programming DLS poses substantial challenges, which stem from their \emph{inherent high parallelism}, \emph{intricate memory hierarchies}, and \emph{specialized instruction sets}. 
First, DLS typically follows parallel programming models, like SIMT or multi-core, diverging from conventional serial programming. This paradigm shift necessitates a deep understanding of task and data parallelism from programmers.
Furthermore, DLS are designed with complex on-chip memory hierarchies to process diverse data types efficiently. For instance, Cambricon MLU uses separate neuron and weight storage for different memory types (i.e., NRAM and WRAM space), while NVIDIA GPU utilizes various memory spaces (e.g., local and shared memory) for multi-level dataflow. Such designs require programmers to explicitly manage complicated memory hierarchies.
Additionally, to suit deep learning's computational demands, DLS often incorporate specialized instruction sets. Examples include Intel's DL Boost VNNI for specific operations and Cambricon's tensor intrinsics 
These specialized intrinsics often come with intricate constraints, adding extra complexity to programming. 
Table~\ref{tab:my_label} illustrates the stated programming difficulty for various DLS.

\subsubsection{Search-based Program Synthesis}

Researchers have studied how to use search-based program synthesis to transform legacy code to high-performance code with minimal manual efforts~\cite{ikarashi2022exocompilation,bansal2023mosaic,vanhattumvectorization, bhatia_et_al:LIPIcs.ECOOP.2023.38}.
The idea is to obtain programs that satisfy semantic correctness by searching through a pre-defined domain-specific language (DSL) or using an SMT solver for constraint solving.
The main advantage of search-based program synthesis is to ensure
the semantic equivalence between the original code and the translated code.
However, these approaches always struggle with large search spaces, which means that they can only be applied to pre-defined DSL and thus cannot generate code for commodity deep learning systems. Additionally, they can only handle computational instructions and are unable to meet optimization requirements for hierarchical storage and parallel capabilities. Thus, the source-to-source translation across different deep learning systems remains an open problem.

\subsubsection{LLM-assisted Program Generation}
LLMs such as Codex~\cite{chen2021evaluating} and StarCoder~\cite{li2023starcoder} are being increasingly used to assist in programming. They can generate programs based on problem descriptions and generate possible code snippets based on the provided code context, helping programmers write code quickly. They also allow programmers to describe problems in natural language without following specific programming language syntax.
However, these LLMs have several limitations. Firstly, they lack a deep understanding of program semantics and cannot comprehend the intent of the code. This means that the generated code may be incorrect or incomplete, requiring further adjustment and optimization. Secondly, these models perform poorly in specific domains or complex problems, because it is typically difficult to obtain high-quality and diverse training data for these problems.

\vspace{-5pt}
\subsection{Motivation}

To understand the design principle of \name{}, we evaluate transcompilation of real-world tensor programs by using state-of-the-art LLM (i.e., GPT-4) and program synthesis tools.

\begin{figure}[!ht]
    \vspace{-10pt}
    \centering
    \includegraphics[width=1.0\linewidth]{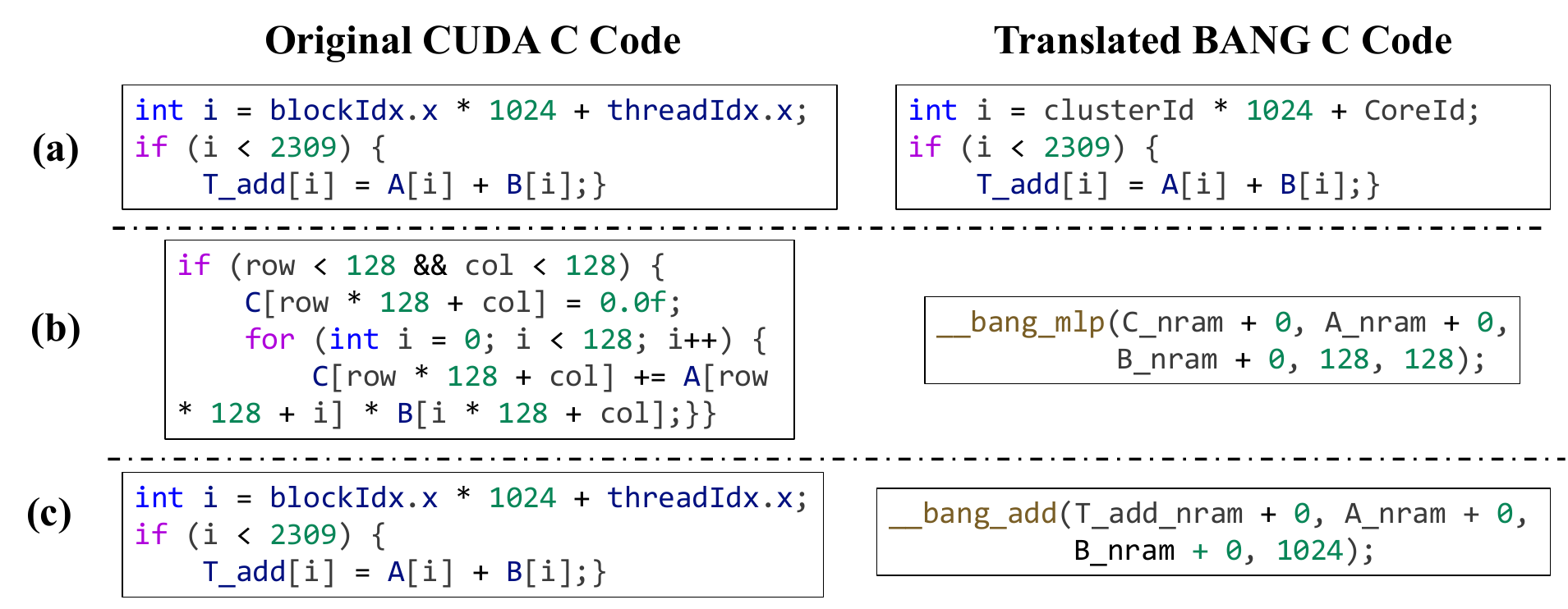}
    \vspace{-20pt}
    \caption{\small The unsuccessful transcompilation examples of GPT-4: (a) Parallelism-related. (b) Memory-related. (c) Instruction-related.}
    \label{fig:error-examples}
    \vspace{-10pt}
\end{figure}

\textbf{Taxonomy of the transcompilation errors}. We first classified all errors introduced in the transcompilation process into 3 categories, each corresponding to specific characteristics of DLS: 
(1) \emph{Parallelism-related errors}, where the transcompilation fails to analyze the loop semantics and generates the incorrect loops or builtin variables of DLS,
(2) \emph{Memory-related errors}, where the transcompilation cannot deal with the intricate memory hierarchies for the DLS, resulting in incorrect memory declarations and usage, and 
(3) \emph{Instruction-related errors}, which occur when the translated code utilizes incorrect instructions or parameters that cannot perform the same computation. Figure~\ref{fig:error-examples} illustrates examples of different errors when transcompiling from CUDA C to BANG C code by GPT-4.
In Figure~\ref{fig:error-examples}(a), the translated code failed to perform the appropriate address offset calculations according to the parallelism characteristics of MLU.
Figure~\ref{fig:error-examples}(b) fails to correctly place the B Tensor in the original GEMM operation into the appropriate memory hierarchy of the MLU (i.e., WRAM).
Figure~\ref{fig:error-examples}(c) tends to replace the original SIMT-based scalar operations with SIMD-based tensorized instructions, but the parameter that indicates the tensor length should be $2309$ rather than $1024$.

Based on the transcompilation error taxonomy, we examine the existing methods and draw the following observations.

\textbf{Observation \#1:} \emph{The single-step LLM-based transcompilation exhibits significant error rates across all three categories, posing challenges for accurately transcompiling programs across deep learning systems.}

Table~\ref{tab::outcome} presents the detailed results for transcompiling CUDA C code to BANG C for real-world operators (detailed in Sec.\ref{sec:evaluation}) using GPT-4. For the zero-shot, the compilation error rate was 100\%, primarily because the LLM struggled with the \emph{complex memory hierarchy} (100\%) and \emph{special instructions} (100\%) due to its insufficient training on DLS-specific datasets. For the few-shot which provided several examples in prompts, although the translated code compiled successfully, it showed functional errors, particularly with parallelism (97.2\%) and instruction-related issues (94.4\%), resulting in an overall computation error rate of 92.3\%. These findings highlight the limitations of single-step LLM-based transcompilations, even with the most advanced LLMs.

\begin{table}[!ht]
\vspace{-5pt}
    \caption{\small Breakdown of the unsuccessful transcompilations produced by GPT-4 based on outcome.(\%)}
    \label{tab::outcome}
    \centering
    \vspace{-5pt}
    \scriptsize
    \setlength{\tabcolsep}{1pt}
    \resizebox{.99\columnwidth}{!}{
  \begin{tabular}{c|ccc|c|ccc|c|}
   \toprule
  \multirow{2}{*}{\textbf{Errors}} & 
    \multicolumn{3}{c|}{\textbf{Compilation}} & \multirow{2}{*}{\textbf{Total}} & \multicolumn{3}{c|}{\textbf{Computation}} & \multirow{2}{*}{\textbf{Total}}\\ \cmidrule{2-4} \cmidrule{6-8}  
     & \textbf{Parallelism} & \textbf{Memory} & \textbf{Instruction} & & \textbf{Parallelism} & \textbf{Memory} & \textbf{Instruction} &  \\  \midrule
    \textbf{Zero-Shot} & 3 & 100 & 100 & 100 & -- & -- & -- & -- \\ 
    \textbf{Few-Shot} & 2.3 & 27.1 & 76.5 & 49.4 & 97.2 & 2.8 & 94.4 & 92.3 \\

\bottomrule 

\end{tabular}
}
\vspace{-5pt}
\end{table}

\begin{figure*}[t]
\vspace{-20pt}
\centering
\includegraphics[width=0.75\linewidth]{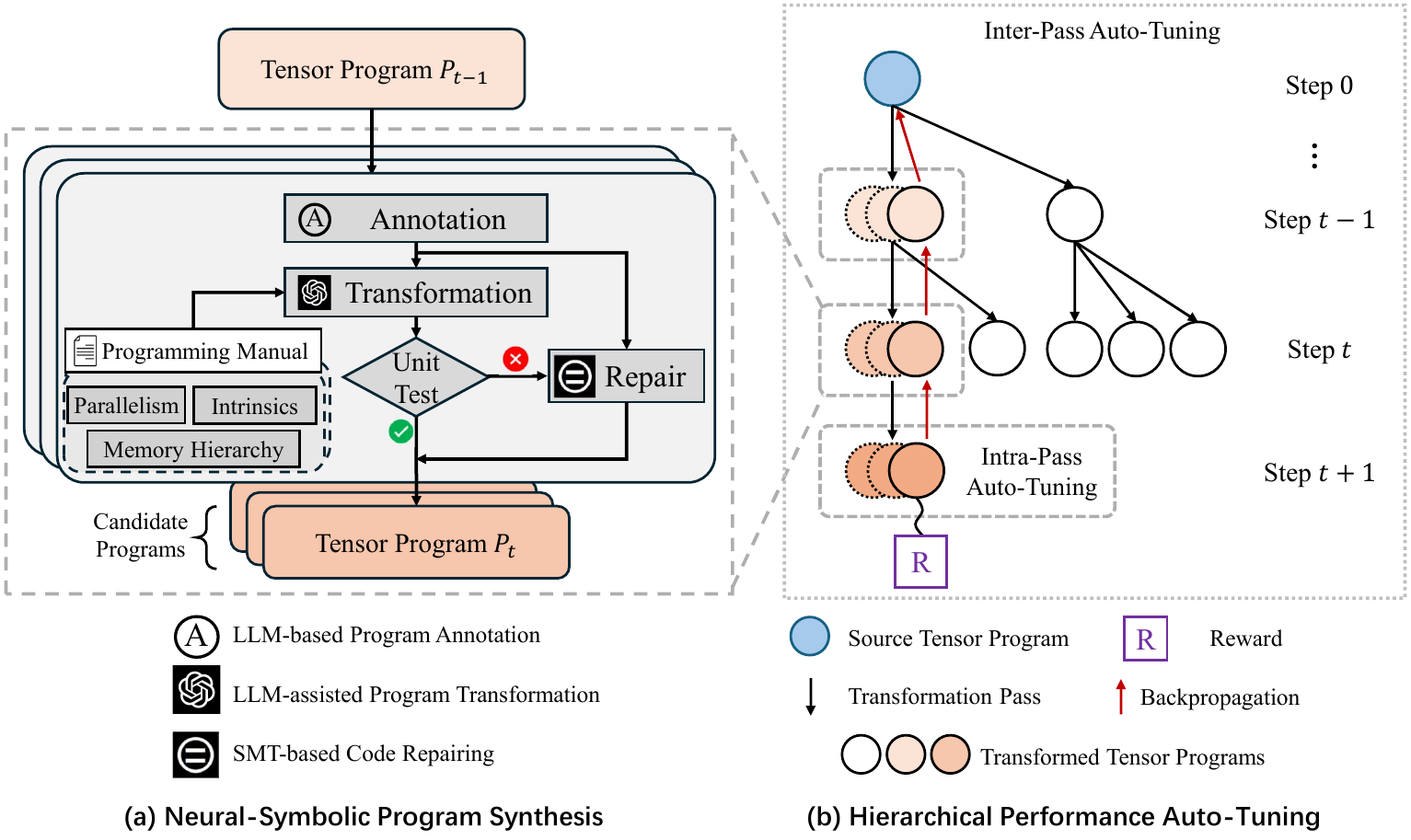}
    \vspace{-10pt}
    \caption{\small\todo{The overview of \name{}, a novel transcompiler for automatic transcompilation of tensor programs across different programming models. The transcompiler consists of two parts: (a) neural-symbolic program synthesis, which utilizes LLM to transform code and repair incorrect transformation through symbolic synthesis with limited scales, and (b) hierarchical performance auto-tuning, which systemically explores both the parameters and sequences of transformation passes.
    }}
    \label{fig:overview}
    \vspace{-15pt}
\end{figure*}

\textbf{Observation \#2:} \emph{Search-based program synthesis excels at optimizing loop bounds and indexing but struggles with high-level program sketches generation, while LLMs are proficient at generating high-level program sketches but prone to errors in low-level details such as loop bounds and indexing.}

\begin{table}[!ht]
\vspace{-5pt}
    \caption{\footnotesize Taxonomy of the search-based program synthesis approach.}
    \label{tab::fuck}
    \centering
    \vspace{-5pt}
    \scriptsize
    \setlength{\tabcolsep}{1pt}
    \resizebox{.99\columnwidth}{!}{
\begin{tabular}{c|c|c|c}
\toprule
                           & \textbf{Method}                        & \textbf{Inputs} & \textbf{Solving Time} \\ \midrule
High-level Program Sketchs & Verified Lifting (Tenspiler~\cite{DBLP:conf/ecoop/QiuCBHSC24}) & Custom IR               & +++                   \\ 
Low-level Program Details  & SMT Solver (Z3~\cite{Z3Solver})                   & SMT Query               & +                     \\ \bottomrule
\end{tabular}
}
\vspace{-10pt}
\end{table}

We analyze the strengths and limitations of search-based program synthesis and LLMs in tensor program transcompilation, highlighting their complementary roles.
Search-based synthesis, as shown in Table~\ref{tab::fuck}, excels at synthesizing and verifying low-level details such as loop bounds and indexing, leveraging its ability to handle mathematical constraints. However, it struggles with generating high-level structures like control flow and efficiently utilizing the memory hierarchy and specialized intrinsics. In contrast, LLMs are highly effective at generating high-level program skeletons but prone to errors in low-level details, such as loop bounds and indexing, as illustrated in Figure~\ref{fig:error-examples}(c).
By combining both approaches, LLMs can generate the high-level program structure, while search-based synthesis refines the low-level details, leading to more accurate, efficient, and reliable program synthesis.

These observations motivate us to decompose the program translation process into a series of LLM-assisted transformations, where program synthesis tools can further enhance the correctness of transcompilation.

\section{Overview}

\begin{table}[t]
    \caption{\footnotesize The transformation passes and their description.}
    \label{tab:table-passes}
    \centering
    \scriptsize
    \vspace{-5pt}
\begin{tabular}{ccc}
\toprule
\textbf{}                     & \textbf{Name}    & \multicolumn{1}{c}{\textbf{Decription}}                                                                          \\ \hline
\multirow{7}{*}{\textbf{(1)}} & Loop Recovery    & Convert parallel variables to sequential for loops                                                               \\
                              & Loop Bind        & Assign a sequential loop to parallel variables                                                                   \\
                              & Loop Split       & Divide a loop into several sub-loops                                                                             \\
                              & Loop Fuse        & Merge several loops into a hyper-loop                                                                            \\
                              & Loop Reorder     & Change the execution orders of loops                                                                             \\
                              & Loop Expansion   & Split a loop body into several loop bodies                                                                       \\
                              & Loop Contraction & Merge the producer in the loop body of consumer                                                                  \\ \hline
\multirow{2}{*}{\textbf{(2)}} & Cache            & \begin{tabular}[c]{@{}l@{}}Adapt to the memory hierarchy for efficient\\ ~~~~~~~~~~~~~~~~~~~ load/store inputs/outputs\end{tabular} \\
                              & Pipeline         & Pipeline of data load/store and computation                                                                      \\ \hline
\multirow{2}{*}{\textbf{(3)}} & Tensorize        & \begin{tabular}[c]{@{}l@{}}Replace a specific loop body to leverage \\ ~~~~~~~~~~~~~~~~~~special intrinsics\end{tabular}           \\
                              & Detensorize      & Restore a specific loop body from special intrinsics                                                             \\ \bottomrule
\end{tabular}
\end{table}

To automatically translate tensor programs across DLS with a correctness guarantee, as shown in Figure~\ref{fig:overview}, \name{} 
consists of two parts, i.e., \textit{Neural-Symbolic Program Synthesis} and \textit{Hierarchical Performance Auto-Tuning}.

\textbf{Neural-Symbolic Program Synthesis}. 
\name{} first decomposes the entire transcompilation process into a series of LLM-assisted transformation passes.
Among these, the 11 passes as listed in Table~\ref{tab:table-passes} can be categorized into 3 classes, i.e., (1)\emph{sequentialization/parallelization}, (2)\emph{memory conversion}, and (3)\emph{(de)tensorization}. 
Specifically, the sequentialization/parallelization passes directly convert a parallel program into its sequential counterpart (e.g., from CUDA C to C) or vice versa, by appropriately mapping built-in parallel variables (e.g., \texttt{threadIdx.x} in CUDA C) to indexing variables. The memory conversion passes bridge the semantic gap between the memory hierarchies of different systems by employing suitable strategies for data movement and access patterns. The (de)tensorization passes either convert sequential code into its parallel equivalent by invoking corresponding tensor intrinsics, automatically retrieved from the target system’s programming manuals, or restore the original sequential code from the tensor intrinsics.
Overall, these passes, each addressing a key characteristic of different DLS—such as parallelism, memory hierarchy, and specialized instructions—are sufficient to perform tensor program transcompilation across various DLS.

To achieve both flexibility and correctness, in each program transformation pass, \name{} adopts a neural-symbolic approach, i.e., utilizing LLM to transform code and repair incorrect transformation through symbolic synthesis with limited scales.
In each transformation pass, the LLM first generates transformed code, and this transformed code is then validated by unit tests; if it fails, it is repaired using small-scale SMT-based symbolic synthesis. We will detail this in Section~\ref{sec:transformation}.

\textbf{Hierarchical Performance Auto-Tuning}. To achieve automatic transcompilation and maximize the performance of transformed programs, \name{} employs a hierarchical auto-tuning approach
to systemically explore both the parameters (intra-pass auto-tuning) and sequences (inter-pass auto-tuning)
of transformation passes.
Specifically, the intra-pass auto-tuning utilizes brute-force search to identify the optimal parameters (e.g., tiling sizes) for program transformation. 
The inter-pass auto-tuning, on the other hand, leverages MCTS to automatically discover both functionally correct and optimal sequences of transformation passes for translating tensor programs across different DLS.
Concretely, the MCTS in the inter-pass auto-tuning consists of four components: tensor programs, transformation pass, backpropagation, and reward. 
At each step \( t-1 \), the MCTS selects a candidate node (i.e., tensor program \( P_{t-1} \)) based on its score and expands the node by transforming it with an available transformation pass, resulting in tensor program \( P_{t} \).
Then, by executing \( P_{t} \), a reward related to execution time is obtained. 
This reward is backpropagated to all ancestor nodes along the expansion path, updating their scores for the selection in the next step \( t \).
We will detail this in Section~\ref{sec:auto-tuning}.

\begin{figure*}[t]
\vspace{-20pt}
\centering
\includegraphics[width=0.95\linewidth]{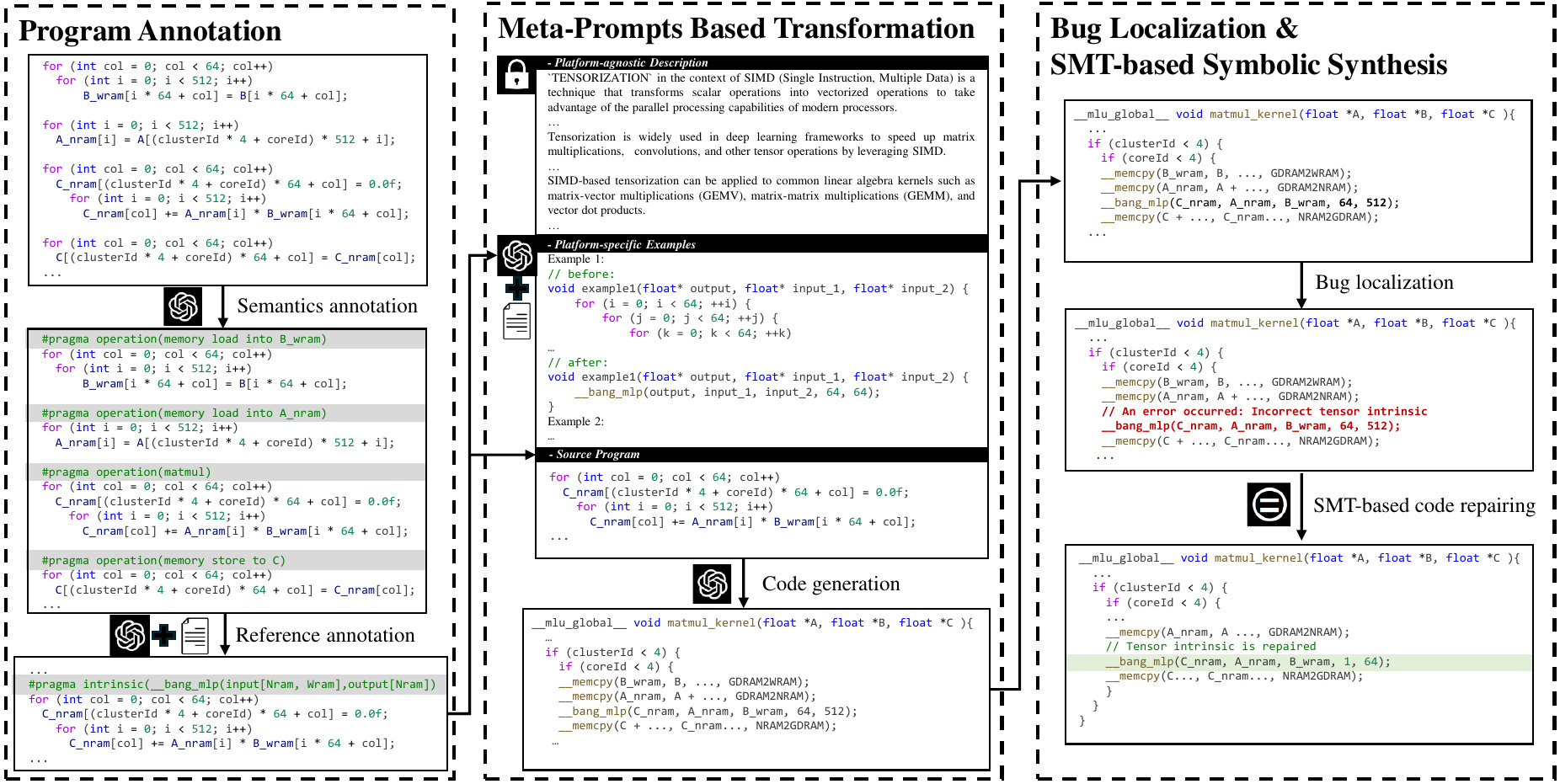}
    \vspace{-10pt}
    \caption{An illustrative example of the proposed neural-symbolic program synthesis on a tensorization case.}
    \label{fig:transformation-example}
    \vspace{-10pt}
\end{figure*}

\vspace{-10pt}
\section{Neural-Symbolic Program Synthesis}\label{sec:transformation}
\name{} breaks down the traditional program transformation process into multiple transformation passes, with each pass covering a full workflow of neural-symbolic program synthesis containing program annotation, meta-prompts based transformation, bug localization, and SMT-based code repairing.
The neural processes, program annotation, and meta-prompts based transformation, leverage the programming manual and the prior knowledge within the LLM to enhance the flexibility of program transformation, enabling \name{} to handle larger-scale code compared to traditional methods. 
Meanwhile, the symbolic processes, bug localization, and SMT-based code repairing, ensure the correctness of the transformed program.
These four processes are illustrated in Figure~\ref{fig:transformation-example}.

\vspace{-10pt}
\subsection{Program Annotation}
For a specific transformation pass, the program annotation process marks code blocks with semantics related to the transformation pass, assisting in the subsequent program transformation process.

As shown in Algorithm~\ref{alg:code_annotation}, program annotation consists of two parts. The first part (lines 2-3) is semantics annotation which involves identifying computational operations with semantics (e.g., \texttt{operation(matmul)}). This part is performed by an LLM. 
The second part (lines 4-11) is reference annotation which involves performing an information retrieval in the programming manual based on each identified computational operation to obtain the corresponding target operation and parameters (e.g., \texttt{intrinsic(\_\_bang\_mlp(input[Nram, Wram], output[Nram])}). This part is first performed by a BM25 search engine~\cite{10.1145/2682862.2682863} which can retrieve related information from the programming manual. Then, the retrieved information and the source program are sent to an LLM to annotate the operator with references. 

Program annotation serves two main purposes. First, by identifying the computation operations in the source program, semantics annotation abstracts the platform-agnostic functional semantics of the source program instead of focusing on platform-specific details, which can help LLMs generate semantically correct programs in the subsequent LLM-assisted program transformation. 
Second, by applying reference annotation, the identified computational operations and their associated hardware characteristics (e.g., memory or computation shape constraints for special intrinsics) can be used in the subsequent meta-prompts based transformation process, aiding LLMs in generating more accurate programs.
\vspace{-10pt}

\begin{algorithm}[!h]
\scriptsize
\SetAlgoLined
\SetKwInOut{Input}{Input}
\SetKwInOut{Output}{Output}
\Input{A source program $P$}
\Output{An annotated program}
\textbf{Given:} LLM, BM25 search engine (SE), programming manual $M$

\text{// Annotate program with identified computation}

$P_{d} \gets LLM(P) $

\text{\footnotesize{// Get the list of computation information}}

$L \gets \text{TraverseComputation}(P_{d})$

\For{$n$ in $L$}{
    \text{\footnotesize{// Retrieve the programming manual of each computation}}
    
    $D \gets SE(n, M)$
    
    \text{// Annotate program with memory spaces or tensor intrinsics}
    
    $P_{d} \gets LLM(P_{d}, D)$
}
\textbf{return} $P_{d}$
\caption{Program Annotation Algorithm}
\label{alg:code_annotation}
\end{algorithm}

\vspace{-20pt}
\subsection{Meta-Prompts based Transformation}
Meta-prompts-based transformation uses LLMs to convert the source program into the target program with a correct program sketch. 
In this approach, meta-prompt is a high-level prompt template that can be adapted to different source programs based on the results of the previous program annotation process. The adapted prompt, along with the source program, is then provided as input to the LLMs. 
Each transformation pass has its own meta-prompt. Meta-prompts mainly consist of three parts: platform-agnostic description, platform-specific examples, and tuning knobs.

\textbf{Platform-agnostic description.}
The platform-agnostic part describes the program's functionality and detailed constraints that must be considered during implementation. For example, in the case of Figure~\ref{fig:transformation-example}, the platform-agnostic part refers to the functionality of the tensorization in the context of SIMD and the application scenarios like deep learning frameworks and common linear algebra kernels. This part of the prompt remains the same across different platforms.

 \textbf{Platform-specific examples.}
Although operators on different platforms share similar functional semantics, their implementation details, e.g., function names and parameters, can vary significantly. As a result, directly transforming programs across platforms with LLMs often leads to incorrect programs.
To address this, we utilize the annotated source program and the LLM to search for platform-specific implementation examples in the target platform's programming manual that are functionally related to the source program. These examples are incorporated into the meta-prompt as the platform-specific component, bridging the gap between the high-level functional semantics of the operators and target platform. This guides LLM in generating the target program.
For instance, in Figure~\ref{fig:transformation-example}, the search results include an example of how to transcompile a \texttt{matmul} from C to BANGC.

\textbf{Tuning knobs.}
The loop split pass and loop reorder pass require more constraints related to the target platform.
This information is included in the meta-prompt optionally.
For example, loop split needs to determine the loop split alignment size for the target platform and use this as the minimum unit to expand into a corresponding search space, which is then used for the auto-tuning of the generated program. Please refer to details in Section~\ref{sec:auto-tuning}.

Based on the meta-prompt, the LLM can generate the transformed code with a generally correct program sketch but error-prone values. Then, we fix these buggy values with symbolic approaches including bug localization and SMT-based code repairing, which will be introduced as follows.

\subsection{Bug Localization}
The feasibility of \name{} relies on reducing the problem scale for code repair, which requires accurately localizing buggy code snippets during unit tests.

For bug localization, the transformed code is validated by the provided unit tests and if it fails, 
Algorithm \ref{alg:error_localization} will be employed to locate the buggy code snippets with unit tests precisely. Initially, we identify the code snippet to be transformed based on the loop index or buffer name and compare the buggy transformed snippet with the original. By inserting print statements after relevant memory locations using binary search, we compare printed values to narrow down the scope of the error-containing snippet. Mappings between source-level control flow statements and diverging inputs then help determine when and where the transcompiled program starts exhibiting erroneous behavior. \name{} subsequently enumerates the constraints in the snippet for each assigned concrete index variable, loop boundary, buffer size, or tensor semantic, and generates queries for the subsequent SMT-based code-repairing process. 

\begin{algorithm}[h]
\scriptsize
\SetAlgoLined
\DontPrintSemicolon
\SetKwInOut{Input}{Input}
\SetKwInOut{Output}{Output}

\Input{Source program $S$, Buggy transcompiled program $E$, Loop index $I$, Buffer name $N$, control-flow matching model $M$}
\Output{Error Nodes $M_{t}$}
$S_{n} \gets \text{AST}(S)$ \; 
$S_{s} \gets \text{ExtractASTSegment}(S_{n}, I, N)$ \;
\text{// Obtain the buffer list} \;
$B_{s} \gets \text{TraverseBuffers}(S_{s})$ \;
$E_{n} \gets \text{BinarySearchErrorNodes}(B_{s}, E, S)$ \;
$M_{b} \gets \text{FindMatchingCodeBlocks}(M, S_{s}, E_{n})$ \; 
\For {$M_{s} \leftrightarrow M_{t} $ in $M_{b}$}{
   \If {$\text{Diverses}(M_{s}, M_{t})$} {
    \KwRet $M_{t}$ 
   } 
}
\caption{Bug Localization Algorithm}
\label{alg:error_localization}
\end{algorithm}

\subsection{SMT-based Code Repairing}
Once the bug is located, the buggy code snippet will be repaired by an SMT solver. 
In the SMT-based code repairing process, we focus on the implementation details such as loop boundaries, indexing, and instruction parameters~\cite{DBLP:conf/ecoop/QiuCBHSC24}, which are orthogonal to the transformation pass.

\begin{algorithm}[h]
\scriptsize
\SetAlgoLined
\caption{SMT-based Code Repairing Algorithm}
\label{alg:code_repair}
\SetKwInOut{Input}{Input}
\SetKwInOut{Output}{Output}
\Input{A source program $P$, error program $E$}
\Output{A repaired program $F$}
\textbf{Given:} Error code snippet $S$ 

$T_{s} \gets \text{AST}(P)$; $S_{s} \gets \text{extract}(T_{s}, S)$

    $T_{e} \gets \text{AST}(E)$; $S_{e} \gets \text{extract}(T_{e}, S)$

    \text{\footnotesize{// Generate the sketch according to the transformation definition.}}
    
    $S_{k} \gets \text{GenerateCodeSketch}(S_{s}, S_{e})$

    \text{\footnotesize{// Generate the smt query}}
    
    $Q \gets \text{CreateSMTQuery}(S_{s}, S_{e})$

    \text{\footnotesize{// Generate the verified code snippet}}

    $R \gets \text{SynthesisCode}(Q)$

$F \gets \text{StitchBack}(P, R, S)$

\textbf{return} $F$

\end{algorithm}

\begin{figure}[t]
    \vspace{-5pt}
    \centering
    \includegraphics[width=1.0\linewidth]{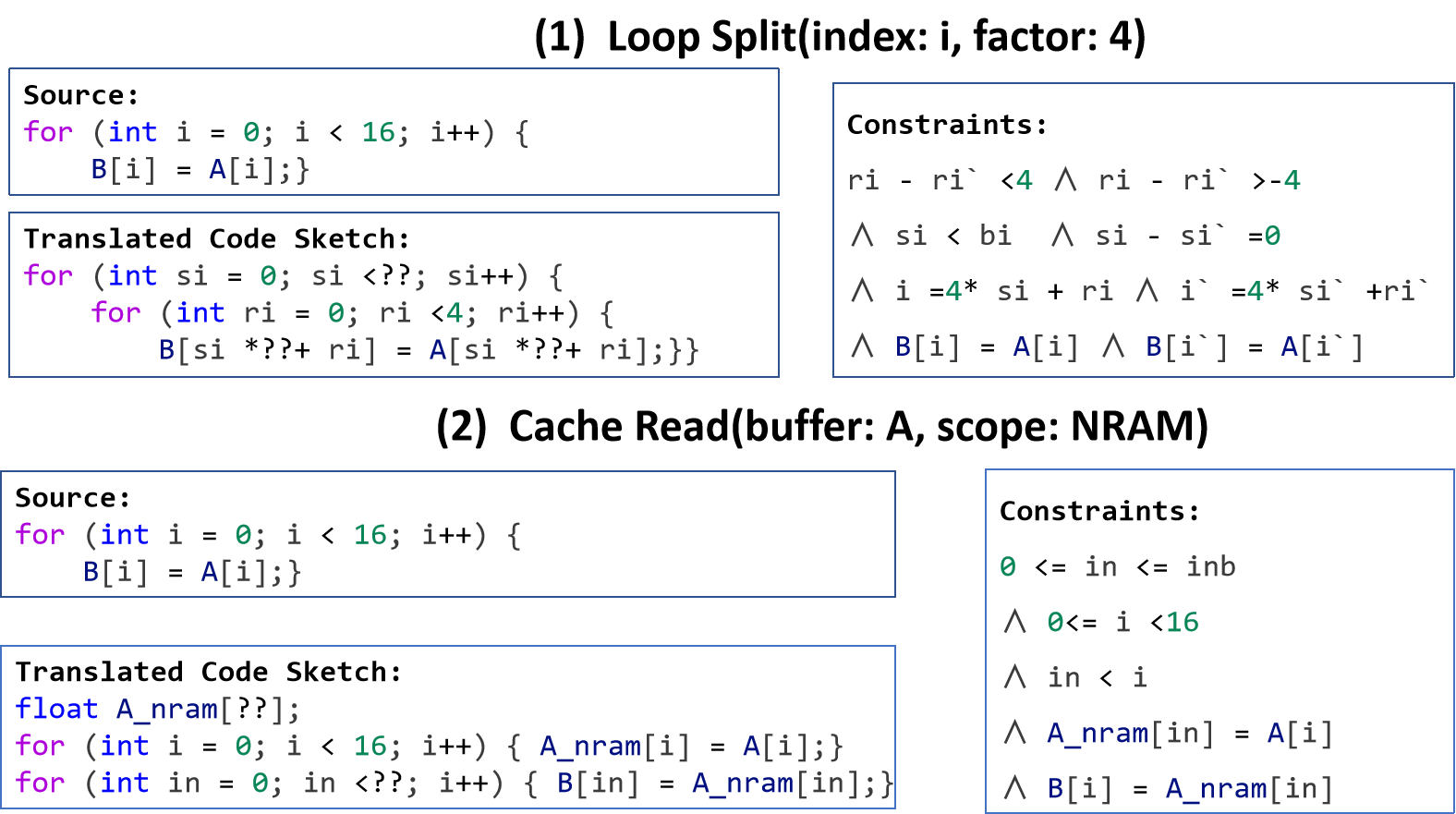}
    \vspace{-15pt}
    \caption{\small The SMT constraints for loop split and cache read.}
    \label{fig:constraints}
    \vspace{-10pt}
\end{figure}

Specifically, as illustrated in Algorithm \ref{alg:code_repair}, \name{} first parses the source code and generates the corresponding sketch based on the definition of the specific loop transformation. Then, the transcompiler explicitly enumerates the corresponding constraints over each assigned concrete index variable and loop boundary, generating code targeting the SMT solver. Finally, by formulating these boundary constraints and buffer index constraints for the SMT solver, the missing expressions in the sketch are filled in. Examples of the constraints for loop split and cache reads are shown in Figure \ref{fig:constraints}.

By symbolically aligning these details with the source program with an SMT solver, we correct errors in the target program transformed by the LLM. This program is outputted as the final program of one transformation pass.

\section{Hierarchical Performance Auto-Tuning}
\label{sec:auto-tuning}
To fully automate code transcompilation and achieve higher performance, \name{} employs a hierarchical auto-tuning approach, which effectively decouples high-level code structures from low-level details, enabling flexible enumeration of high-level program sketches (i.e., strategizing parallelization and thread bindings), as well as efficient sampling of low-level specifics (i.e., tiling size and loop orders). As a result, transcompilation exploration space can be formalized as follows:

\vspace{-5pt}
\small\begin{equation}
S = \left\{ S^{(n)} \mid 
\begin{array}{l}
S^{(t)} = \text{apply}(\text{apply}(S^{(t-1)}, d_t), k_t), \\
\forall d_t \in D_t, 1 \leq i \leq n, \\
\forall k_t \in K_t, 1 \leq t \leq n
\end{array}
\right\}
\label{eq:searchspace}
\end{equation}
\normalsize
where $s^{0}$ denotes the source tensor program, and $d_{t}$ represents a random selected pass from the pre-defined passes $D_{t}$, $k_{t}$ represents a tuning option from the pre-defined tuning knobs $K_{t}$, and $n$ is number of transformation. Thus, the size of the exploration space aligns with the number of
pass sequences and tuning knobs. We have:
\small\begin{equation}
|S| = |D_{1}| \times |K_{1}| \times |D_{2}| \times |K_{2}| \times \ldots \times |D_{n}| \times |K_n|
\end{equation}
\normalsize
To explore this huge search space, \name{} puts forward a hierarchical performance auto-tuning consisting of two main components creatively: (1) \textbf{Intra-Pass Auto-Tuning}, which focuses on fine-tuning the performance of translated programs; and (2) \textbf{Inter-Pass Auto-Tuning with MCTS}, which is responsible for constructing a vast valid search space to identify the most optimal code transcompilation.

\subsection{Intra-Pass Auto-Tuning}
During the transformation process, passes such as \texttt{loop split} and \texttt{loop reorder} are tasked with making a series of critical decisions, including the split size and loop order, which significantly affect the performance of the resulting tensor program. To address these challenges, we consider implementing auto-tuning, an effective technique for automatically generating high-performance programs. Specifically, auto-tuning is implemented as a functional module that interacts with the aforementioned passes to create a search space consisting of multiple candidate programs. It then explores this space to identify the optimal program configuration.

\begin{figure}[t]
    \vspace{-5pt}
    \centering
    \includegraphics[width=1.0\linewidth]{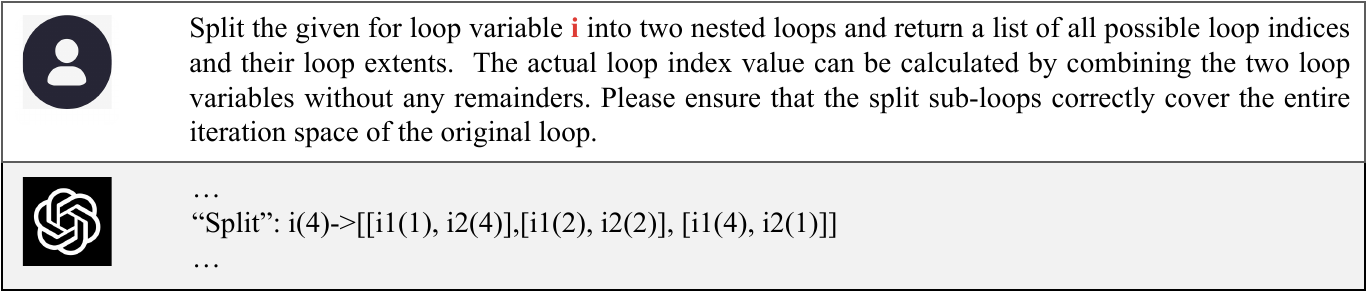}
    \vspace{-20pt}
    \caption{\small The auto-tuning prompt for loop split.}
    \label{fig:prompt-loopsplitreorder}
    \vspace{-15pt}
\end{figure}

\textbf{Search space generation.} The basis of intra-pass auto-tuning is to construct a search space with a large number of candidate programs, and our approach generates such space with the help of LLM and programming manuals. As shown in Figure~\ref{fig:overview}, the auto-tuning module first interacts with the \emph{loop split} to generate multiple programs with different adjustable parameters including: (1) the number of blocks for each logical loop, (2) the loop order after making decisions on (1). This is achieved by using specially designed meta-prompts as shown in Figure~\ref{fig:prompt-loopsplitreorder}, which are then transferred to the prompts of the loop split and loop reorder pass. Then, the intra-pass auto-tuning module interacts with other passes to generate different program sketches, such as different loop binding strategies.

\textbf{Search space exploration.} Depending on the characteristics of the target code, e.g., in the case of BANG C language with larger instruction granularity resulting in a small search space, we directly employ a brute-force search approach for exploring the program with optimal performance. Actually, the size of the search space varies significantly for different DLS. Take the Matmul operation(512 × 512 × 512) as an example, on GPU the size of search space (K in Equation \ref{eq:searchspace}) is $150$ while on MLU the size is only $10$.

\vspace{-5pt}
\subsection{Inter-Pass Auto-Tuning with MCTS}
We employs MCTS, which formulates the transcompilation as a Markov decision process where each intermediate program is represented as a state and the actions are the different passes that could be applied next. The reward is proportional to the execution time improvement. The solution would be the actions that lead to the optimal pass sequence.

\textbf{Reward function.} To enhance the accuracy of the cost model while minimizing additional overhead, we add real execution time measurements each time a new transformation pass is declared. While the program is not fully translated, MCTS runs in parallel with the current programs, records the best real throughput as the reward of the current transformation sequence and its corresponding program.
At step $t$, the reward function is

\vspace{-10pt}

\footnotesize
\begin{align}
T_{ti} &=
\begin{cases}
    T(p_{ti}), & \text{if } p_{ti} \\[5pt]
    0, & \text{otherwise}
\end{cases} \\
R_{t} &= \max(T_{ti})
\end{align}
\vspace{-10pt}
\normalsize

where $p_{t}$ denotes the transformed programs at iteration $t$, $i$ is the index of transformed program within intra-pass auto-tuning search space, and $T$ is the throughput of program. 

\textbf{Searching parameters setting.}
MCTS may fail to effectively reach a terminal node in a given iteration or even fail to generate the final transcompiled program. Therefore, to determine the appropriate maximum search depth(i.e., $n$ in Equation \ref{eq:searchspace}) and the total number of simulations, we conducted design space exploration to balance search time and reward. Specifically, the maximum search depth of MCTS should exceed 11 passes, as all passes are crucial and may be repeated, with the number of simulations directly impacting both search time and reward. Ultimately, we selected $N=13$ and 512 simulations for MCTS, which provides optimal performance of the transcompiled program within a reasonable search time (e.g., hours).
\vspace{-5pt}

\section{Evaluation Methodology}
\label{sec:evaluation}
\vspace{-5pt}

\textbf{Evaluated Platforms.}
We conduct experiments on $4$ different deep learning systems with their own programming interfaces as follows: 1) \textbf{Intel Gold 6348 CPU with VNNI Extension}, which utilizes a special instruction set for deep learning; 2) \textbf{NVIDIA A100 GPU with CUDA C}, which follows the SIMT programming model; 3) \textbf{AMD MI200 with HIP}, which provides an alternative to CUDA language; and 4) \textbf{Cambricon MLU with BANG C}, which follows SIMD programming model on a DSA.

\textbf{Evaluated Benchmarks.}
We evaluate \name{} on 21 widely-used deep learning operators (shown in Table \ref{tab::benchmark}), which can be grouped into $6$ types of operations including \emph{MatMul}, 
\emph{Convolution}, 
\emph{Activation}, 
\emph{Pooling}, 
\emph{Element-wise}, 
and \emph{LLM} operation. 
Each operator is further evaluated by $8$ typical shapes extracted from real network networks such as GPT~\cite{GPT4}, LLaMA-2~\cite{touvron2023llama}, DAT~\cite{xia2022vision}, BERT~\cite{devlin2018bert}, ResNet~\cite{He16CVPR}, and MobileNet~\cite{howard2017mobilenets}, VGG~\cite{Simonyan15VGG} and thus there are 168 test cases in total for evaluation. The test cases range from $7$ to $214$ lines of code and cover various types and shapes, which can comprehensively assess \name{}'s capabilities.

\begin{table}[!ht]
    \vspace{-5pt}
    \caption{Evaluated Benchmark for \name{}}
    \label{tab::benchmark}
    \centering
    \vspace{-5pt}
    \scriptsize
    \setlength{\tabcolsep}{2pt} 
    \begin{tabular}{@{}c@{}c@{}cccc@{}@{}c}
        \toprule
        \multirow{2}{*}{\textbf{Type}} & \multirow{2}{*}{\textbf{Operators}} & \multicolumn{4}{c}{\textbf{Lines of Code}} & \multirow{2}{*}{\textbf{Cases}}\\
        \cmidrule(lr){3-6}
        & & \textbf{CUDA C} & \textbf{BANG C} & \textbf{Hip} & \textbf{C with VNNI} \\ 
        \midrule
        \multirow{2}{*}{MatMul} 
        & GEMM, GEMV & 26, 12 & 15, 11 & 25, 12 & 41, 34 & \multirow{2}{*}{24}\\
        & Batch GEMM & 29 & 16 & 28 & 47 & \\ 
        \midrule
        \multirow{3}{*}{Convolution} 
        & Conv1D & 10 & 14 & 10 & 9  & \multirow{3}{*}{32}\\
        & Conv2D NHWC, NCHW & 32, 30 & 23, 30 & 32, 30 & 83, 85 &\\
        & Depthwise Conv & 27 & 24 & 27 & 17 &\\ 
        \midrule
        \multirow{2}{*}{Activation}  
        & ReLU, Softmax & 7, 24 & 18, 18 & 7, 24 & 14, 28 & \multirow{2}{*}{32}\\
        & GeLU, Sigmoid & 7, 10 & 29, 30 & 7, 10 & 14, 14 & \\ 
        \midrule
        Elementwise 
        & Add, Sign & 20, 12 & 26, 29 & 18, 12 & 17, 14 & 16\\ 
        \midrule
        \multirow{2}{*}{Pooling} 
        & MaxPool, AvgPool & 25, 30 & 25, 25 & 25, 30 & 31, 34 & \multirow{2}{*}{32}\\
        & MinPool, SumPool & 23, 29 & 25, 25 & 23, 29 & 33, 30 & \\ 
        \midrule
        \multirow{3}{*}{LLM} 
        & LayerNorm & 42 & 35 & 42 & 46 & \multirow{3}{*}{32}
\\
        & Deformable Attention & 139 & 191 & 139 & 214 & \\
        & Self Attention, RMSNorm & 54, 25 & 64, 18 & 54, 25 & 111, 18 &\\ 
        \bottomrule
    \end{tabular}
\end{table}

\textbf{Comparison Baselines}.
The comparison baselines include state-of-the-art LLM-based approaches and rule-based approaches.
Regarding the LLM-based approaches, we select the \textbf{GPT-4 Zero-Shot}, \textbf{GPT-4 Few-Shot}, \textbf{OpenAI o1 Zero-Shot}, and \textbf{OpenAI o1 Few-Shot} as baselines for each transcompilation direction. 
Regarding the rule-based approaches, we compare \name~with two established approaches: \textbf{PPCG}~\cite{10.1145/2400682.2400713} and \textbf{HIPIFY}\cite{HIPIFY}.
PPCG utilizes polyhedral models for auto-parallelization from C to CUDA
HIPIFY is a vendor-provided tool to migrate CUDA code to AMD HIP code.
Note that rule-based approaches are not available for other transcompilation directions, which is primarily due to the complexity and significant human effort required in developing such tools.
Regarding the search-based program synthesis approaches, as they struggle with the large search space problem, they cannot perform transcompilation directly across different DLS.

\textbf{Evaluation  Metrics.}
We use three key metrics to evaluate the performance and accuracy of \name{}:
(1) \textbf{Compilation accuracy} is the ratio of programs that are correctly compiled. This metric gauges \name{}'s ability to successfully pass compiler checks, reflecting its proficiency in processing source code to generate error-free programs.
(2) \textbf{Computation accuracy} is introduced in this paper as a crucial metric that assesses the functional correctness of the translated code, deeming a generated code correct if it passes a set of unit tests.
(3) \textbf{Execution performance} evaluates the performance of the translated code, highlighting the significance and practicality of the proposed method in real-world applications.
In these metrics, we execute the translated code on real platforms and verify its correctness.

\vspace{-5pt}
\section{Experimental Results}

\begin{table*}[t]
  \vspace{-10pt}
  \centering
  \caption{Experimental results on different transcompilation directions. (\%)
 }
  \label{tab:code_translation}
  \vspace{-5pt}
  \scriptsize
  \begin{tabular}{c|c|cccc|cccc}
   \toprule 
   \multirow{3}{*}{\textbf{Source}} & \multirow{3}{*}{\textbf{Method}} &
    \multicolumn{4}{c|}{\textbf{Compilation Accuracy}} & \multicolumn{4}{c}{\textbf{Computation Accuracy}} \\ \cmidrule{3-6} \cmidrule{7-10}  
     &  & \textbf{CUDA C} & \textbf{BANG C} & \textbf{Hip} & \textbf{C with VNNI} & \textbf{CUDA C} & \textbf{BANG C} & \textbf{Hip} & \textbf{C with VNNI} \\ 
     \cmidrule{1-10}
    \multirow{7}{*}{\textbf{CUDA C}} 
    & GPT-4 Zero-Shot  & -- & 0 & 82.7 & 9.5 & -- & 0 & 82.7 & 4.2\\
    & OpenAI o1 Zero-Shot  & -- & 0 & 85.7 & 61.9 & -- & 0 & 82.7 & 60.7\\
     & GPT-4 Few-Shot  & -- &50.6 & 97.0 & 84.5  & -- & 7.7& 96.4 & 30.4\\
     & OpenAI o1 Few-Shot  & -- & 51.8 & 98.2 & 85.1 & -- & 48.2 & 98.2 & 55.4 \\
     & \namewosmt~& -- & 82.7 & 98.2 & 88.1 & -- & 54.2 & 98.2 & 58.3\\
     & \namewosmt~+ Self-Debugging  & -- & 87.5 & 98.8 & 89.3 & -- & 54.8& 98.2& 58.9 \\ 
    & \name & -- & \textbf{100}  & \textbf{100} & \textbf{100} & -- & \textbf{91.7} & \textbf{100} & \textbf{95.2} \\ 
     \midrule
    \multirow{7}{*}{\textbf{BANG C}} 
    & GPT-4 Zero-Shot & 24.4 & -- & 26.8 & 0  & 0 & -- & 0 & 0\\
    & OpenAI o1 Zero-Shot  & 27.4 & -- & 97.0 & 9.5 & 0 &-- & 0 & 4.2\\
     & GPT-4 Few-Shot & 69.0 & -- & 66.1 & 23.8  & 6.5 & -- & 6.5 & 13.1\\
     & OpenAI o1 Few-Shot  & 71.4 & -- & 97.0 & 41.7 & 10.1 & -- & 7.7 & 23.2\\
     & \namewosmt~& 85.1 & -- & 84.5 & 47.6  & 77.4 & -- & 78.6 & 41.1\\
     & \namewosmt~+ Self-Debugging & 88.1 & -- & 88.7 & 50.6  & 77.4 & -- & 78.6 & 41.1 \\ 
      & \name & \textbf{100} & -- & \textbf{100} & \textbf{100} & \textbf{95.8}  & -- & \textbf{97.0} & \textbf{95.2}\\
     \midrule
    \multirow{7}{*}{\textbf{Hip}} 
    & GPT-4 Zero-Shot& 97.0 & 0 & -- & 23.8 & 97.0 & 0 & -- & 5.4\\
    & OpenAI o1 Zero-Shot  & 98.2 & 0 & -- & 45.8 & 98.2 & 0 & -- & 4.2\\
     & GPT-4 Few-Shot & 97.0 & 35.1& -- & 85.1 & 97.0 & 5.4& -- & 24.4\\
     & OpenAI o1 Few-Shot  & 98.8 & 42.3 & -- & 88.7 & 98.2 & 9.0& -- & 30.4\\
     & \namewosmt~& 98.2 & 60.7 & -- & 65.5 & 97.6 & 52.4 & -- & 57.1\\
     & \namewosmt~+ Self-Debugging &  98.8 & 62.5 & -- & 66.1  & 98.2 & 52.4 & -- & 57.1 \\ 
    & \name & \textbf{100} & \textbf{100} & -- & \textbf{100} &\textbf{100} & \textbf{86.9} & -- & \textbf{96.4} \\ 
    \midrule
    \multirow{7}{*}{\textbf{C with  VNNI}} 
    & GPT-4 Zero-Shot & 57.1 & 0 & 60.1 & -- & 8.3 & 0 & 8.9 & --\\
    & OpenAI o1 Zero-Shot  & 66.1 & 0 & 97.0 & -- & 10.1 & 0 & 96.4 & --\\
     & GPT-4 Few-Shot & 81.5 & 41.7 & 74.4 & --  & 14.3 & 6.0 & 12.5 & --\\
     & OpenAI o1 Few-Shot  & 87.5 & 55.4 & 97.0 & -- &51.2 & 10.7 & 96.4 & --\\
     & \namewosmt~& 95.8 & 78.0 & 87.5 & -- & 83.9 & 58.3 & 85.7 & --\\
     & \namewosmt~+ Self-Debugging & 97.0 & 84.5 & 89.3 & -- & 83.9 & 58.3 & 85.7 & -- \\ 
           & \name & \textbf{100} & \textbf{99.4}&\textbf{100}& -- & \textbf{98.2} & \textbf{88.7} & \textbf{99.4} & --\\ 
    \bottomrule 
  \end{tabular}
    \vspace{0pt}
\vspace{-15pt}
\end{table*}

\subsection{Evaluations on Accuracy}
We present the evaluations on compilation/computation accuracy in Table~\ref{tab:code_translation} and Table~\ref{tab:exp-rule}, where we compare \name{} with state-of-the-art methods in different transcompilation directions.
We conclude that
\textbf{(1) \name{} performs the best in all directions with close to 100\% accuracy for compilation and 86.9\% to 100\% accuracy for computation.} This clearly indicates that \name{} is capable of handling source-to-source code translation tasks on various DLS with minimal human efforts, bringing revolutionary advancements to the DLS programming domain.
\textbf{(2) \name{} performs better than the SOTA LLM-based methods} (Table~\ref{tab:code_translation}). Although the LLM-based methods have achieved high accuracy in certain cases, it is challenging for them to reach 100\% accuracy due to the uncertainty of LLMs. This means that LLM-based methods cannot be applied to transcompilers which have an extremely high demand for accuracy. In contrast, our approach can achieve 100\% accuracy in most situations, demonstrating its practical applicability as a transcompiler.
\textbf{(3) \name{} performs better than the SOTA rule-based methods} (Table~\ref{tab:exp-rule}). For C~$\rightarrow$~CUDA C, \name{} achieves 100\% compilation and 98.2\% computation accuracy which is $\sim50\%$ higher than PPCG. For the easier CUDA C~$\rightarrow$~HIP task, \name{} successfully converts and executes with 100\% accuracy, outperforming HIPIFY, which achieves 85.7\%. 
Also, this result shows that \name{}'s flexibility across various DLS without much adaptation cost while rule-based methods cannot.

We conduct an ablation study on \name{}'s SMT solver for further analysis of the neural-symbolic synthesis paradigm, shown in Table~\ref{tab:code_translation}.
Without the SMT solver, although \name{} still achieves better accuracy than LLM-based methods, it cannot reach 100\% accuracy (e.g., 52.4\% computation accuracy in HIP~$\rightarrow$~BANG C direction). This situation persists even after incorporating \name{} with the SOTA LLM-based code generation method, Self-Debugging
\cite{chen2023teaching}. Due to transcompilers' high demand for accuracy, the methods that cannot achieve or close to 100\% accuracy can be hardly applied in practice. 
In contrast, leveraging SMT-based code repairing can significantly improve the computation accuracy of program transcompilation, achieving a range of 86.9\% to 100\%.
This underscores the necessity of the SMT solver or, more broadly, neural-symbolic synthesis for the transcompiler.

\begin{table}[!ht]
    \caption{Accuracy comparison to rule-based methods.}
    \label{tab:exp-rule}
    \vspace{-5pt}
    \centering
        \scriptsize
\begin{tabular}{cccc}
\toprule
\textbf{Direction}                                                                            & \textbf{Method} & \begin{tabular}[c]{@{}c@{}}\textbf{Compilation}(\%)\end{tabular} & \begin{tabular}[c]{@{}c@{}}\textbf{Computation}(\%)\end{tabular} \\ \midrule
\multirow{2}{*}{\begin{tabular}[c]{@{}c@{}}CUDA C~$\rightarrow$~\\ HIP\end{tabular}} & Hipify &   85.7  & 85.7 \\ \cmidrule{2-4} 
                                                                                     & \name  &       100                                                              &  100    \\ \midrule
\multirow{2}{*}{\begin{tabular}[c]{@{}c@{}}C~$\rightarrow$~\\ CUDA C\end{tabular}}   & PPCG   &          47.6                                                        &                              47.6                                \\ \cmidrule{2-4}                                 & \name  &100 &  98.2                                                        \\ \bottomrule
\end{tabular}
\vspace{-15pt}
\end{table}

\vspace{-0pt}

\begin{figure*}[t]
    \vspace{-5pt}
    \centering
    \includegraphics[width=0.85\linewidth]{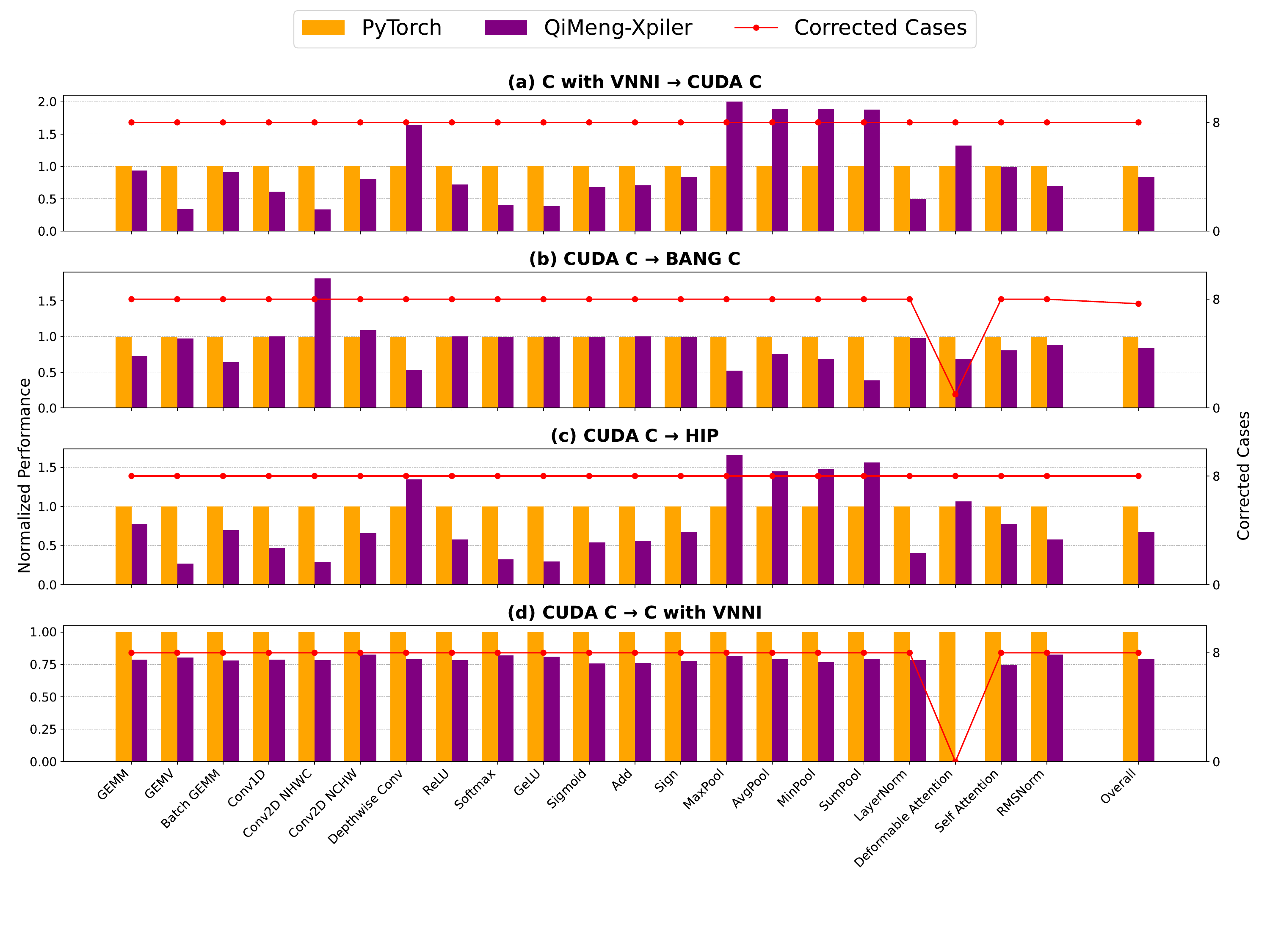}
    \vspace{-20pt}
    \caption{\footnotesize \todo{Performance evaluations on four common transcompilation directions and various operators. The performance of programs generated by \name{} is compared to its manually optimized counterparts, PyTorch, with backend libraries such as cuDNN/cuBLAS, CNNL, rocBLAS, and oneDNN.}}  
    \label{fig:7-3}
    \vspace{-15pt}
\end{figure*}

\subsection{Evaluations on Execution Performance}
We present the performance evaluations in Figure~\ref{fig:7-3}, where we compare the performance of programs generated by \name{} with its manually optimized counterparts (e.g., PyTorch with backend libraries such as cuDNN/cuBLAS, CNNL, rocBLAS, and oneDNN) on the four most common program transcompilation directions, i.e., C with VNNI → CUDA C, CUDA C → BANG C, CUDA C → HIP, and CUDA C → C with VNNI.
Note that we also show the functional correct cases on each type of operator in the line chart and report the average performance across them.

Results show that \name{} achieves an average performance of 0.78$\times$ and up to 2.00$\times$ better than its manually optimized counterparts across the four transcompilation directions and various operators, demonstrating its remarkable transcompilation capabilities for generating high-quality tensor programs.

\vspace{-5pt}
\subsection{Case Study: Trancompilation from CUDA C to BANG C}
\vspace{-5pt}


We further demonstrate the superiority and effectiveness of \name{} by comparing the transcompilation results of different methods in the CUDA C~$\rightarrow$~Bang C transcompilation direction. 
This task is particularly challenging and representative because 1) it involves transcompiling between two distinct programming models, namely translating from SIMT programs to SIMD programs, and 2) it targets an uncommon language, which has significantly less training data available for LLMs.

Overall, \name{} achieves much more improvements in compilation/computation accuracy in translating CUDA C to BANG C compared to other transcompilation directions.
 While OpenAI o1 Zero-Shot shows excellent performance in some transcompilation directions, such as CUDA C~$\rightarrow$~HIP, with compilation and computation accuracy of 85.7\% and 82.7\% respectively, its performance in CUDA~$\rightarrow$~Bang translation is notably lacking, with results at 0\%. 
This shortfall is primarily due to the aforementioned two challenges, which exceeds OpenAI o1 Zero-Shot’s capabilities.
 \begin{figure}[t]
 
    \vspace{-5pt}
    \centering
    \includegraphics[width=0.8\linewidth]{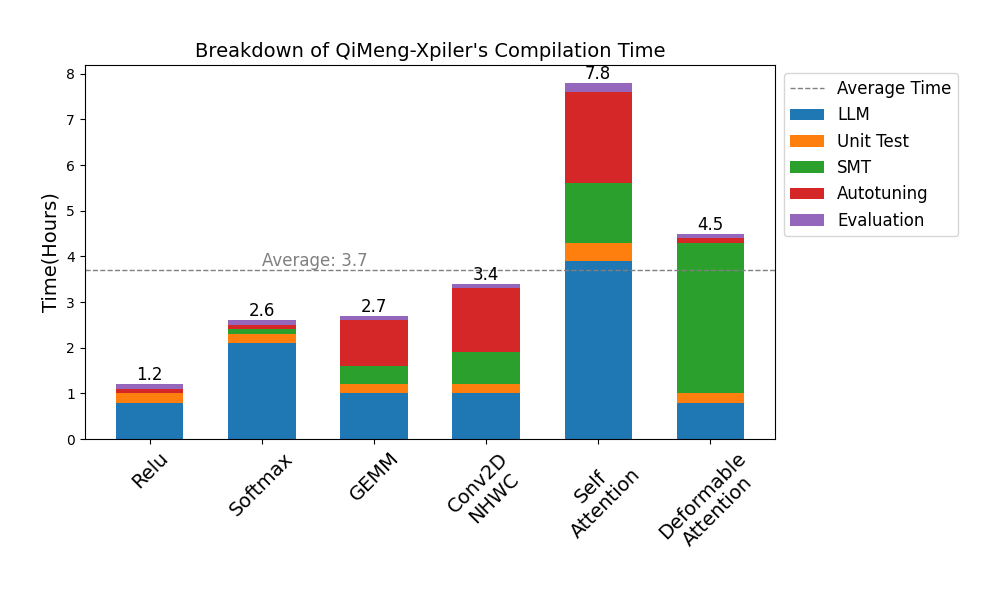}
    \vspace{-20pt}
    \caption{\small Breakdown of \name{}’s compilation time.}
    \label{fig:compilation_time}
\end{figure}

Although OpenAI o1 Few-Shot manages to boost compilation and computation accuracy substantially from 0\% to 50.6\% and 0\% to 7.7\%, respectively, by providing CUDA~$\rightarrow$~Bang translations examples, there still exists a large room attaining the complete accuracy. 
In contrast, \namewosmt~ achieves much higher compilation and computation accuracy of 82.7\% and 54.2\% correspondingly. 
This improvement stems from two key approaches adopted in \namewosmt~:
1) it gains a comprehensive understanding of the target language's keywords, syntax, and programming APIs by automatically annotating programs via document retrieval, thus compensating for the lack of domain-specific knowledge, and 
2) whereas typical transcompilers attempt intricate inter-language translations in a single synthesis step, \namewosmt~splits the entire process into a series of LLM-based transformation passes. 
This enables \namewosmt~to effectively facilitate smooth translations between two substantially different source and target languages.

Crucially, with the integration of SMT methods, \name{} attains remarkable accuracy in the CUDA C~$\rightarrow$~BANG C translation, achieving 100\% in compilation and 91.7\% in computation. This accuracy is largely due to \name{}'s implementation of small-scale symbolic synthesis, which ensures the functional equivalence of each transformation pass guided by the LLM. 
In conclusion, our experiments prove that \name{} successfully accomplishes transcompiler tasks across different DLS, even for a new and relatively uncommon DLS.

\vspace{-15pt}
\subsection{\todo{Compilation Time}}
Figure~\ref{fig:compilation_time} shows the compilation time of 6 typical operators when translating from CUDA C to BANG C, which ranges from \todo{1.2 to 7.8 hours, with 3.7 hours on average}.
We further analyze the breakdown of \name{}'s compilation time and conclude that: 1) SMT is only triggered when the LLM fails to produce a correct translation, so for simpler programs, the proportion of time spent on SMT is smaller; 2) For matrix multiplication-like operators with a larger search space, the proportion of time spent on autotuning significantly increases; 3) Compilation time increases as the number of special intrinsics used in the program grows.

\vspace{-15pt}
\subsection{Productivity Improvement}
We further evaluate how \name{} improves programming productivity in real-world scenarios on $2$ representative DLS, i.e., GPU and MLU.
Specifically, we compare the development costs and performance achieved by manually implementing tensor programs versus those generated by \name{} through automatic transcompilation, on the most challenging operator, Deformable Attention, with $\sim$ 200 LoCs.

We invited two CS master's students and two software engineers as junior and senior coders, respectively. Note that performance is normalized against the manually implemented programs by the senior coders.
Table~\ref{tab::development_cost} shows that, programming productivity is improved by up to 34.3$\times$ for GPU and 96.0$\times$ for MLU via transcompiling legacy tensor program.
Although \name{} fails to automatically generate a functional-correct program for MLU, both junior and senior coders could debug the program with minimal effort, requiring an additional 3 hours and 0.5 hours, respectively. 

\begin{table}[!ht]
    \vspace{-5pt}
    \caption{Productivity Improvement by \name{}.}
    \label{tab::development_cost}
    \centering
    \vspace{-5pt}
    \tiny
\begin{tabular}{cc|cc|cc}
\toprule
\multicolumn{2}{c|}{\multirow{2}{*}{\textbf{\begin{tabular}[c]{@{}c@{}}Deformable Attention\\ ( $\sim$ 200LoCs)\end{tabular}}}} & \multicolumn{2}{c|}{\textbf{\begin{tabular}[c]{@{}c@{}}CUDA C-\textgreater\\ BANG C\end{tabular}}} & \multicolumn{2}{c}{\textbf{\begin{tabular}[c]{@{}c@{}}C with VNNI -\textgreater \\ CUDA C\end{tabular}}} \\ \cmidrule{3-6} 
\multicolumn{2}{c|}{}                                                                                                     & Costs                                        & Performance                                      & Costs                                           & Performance                                         \\ \midrule
\multicolumn{1}{c|}{\multirow{3}{*}{\begin{tabular}[c]{@{}c@{}}Senior\\ Coder\end{tabular}}}         & Manual             & $\sim$6 d                                       & 100\%                                            & $\sim$1 d                                           & 100\%                                               \\
\multicolumn{1}{c|}{}                                                                                & w/ QiMeng-Xpiler          & 4.5 + 0.5 h                                     & 69.20\%                                          & 2.1     h                                           & 132.50 \%                                           \\
\multicolumn{1}{c|}{}                                                                                & Time Saving        & $\sim$28.8$\times$                                       &                                                  & $\sim$11.4$\times$                                           &                                                     \\ \midrule
\multicolumn{1}{c|}{\multirow{3}{*}{\begin{tabular}[c]{@{}c@{}}Junior\\ Coder\end{tabular}}}         & Manual             & $\sim$30 d                                      & 49.85\%                                          & $\sim$3 d                                     & 75.76\%                                             \\
\multicolumn{1}{c|}{}                                                                                & w/ QiMeng-Xpiler          & 4.5 + 3 h                                       & 65.17\%                                          & 2.1   h                                             & 132.50 \%                                           \\
\multicolumn{1}{c|}{}                                                                                & Time Saving        & \textbf{$\sim$96.0$\times$}                                      &                                                  & \textbf{$\sim$34.3$\times$}                                           &                                                     \\ \bottomrule
\end{tabular}
\end{table}

\vspace{-15pt}
\subsection{Failure Case}
\begin{figure}[!ht]
    \vspace{-5pt}
    \centering
    \includegraphics[width=1.0\linewidth]{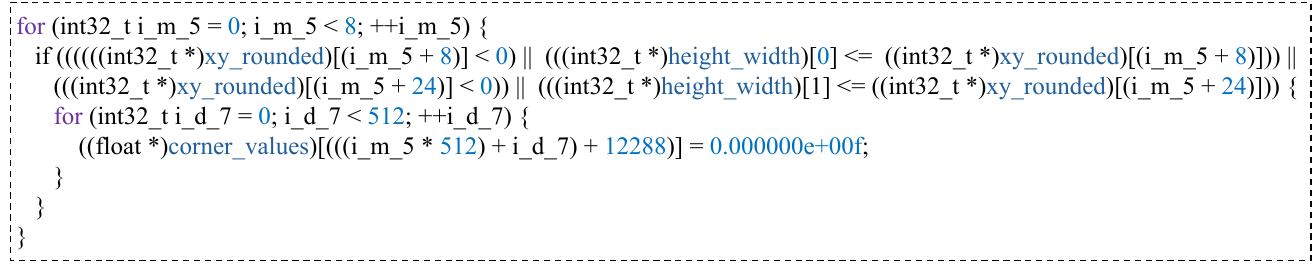}
    \vspace{-20pt}
    \caption{\small The complex control flow in \texttt{Deformable  Attention}.}
    \label{fig:error_translated}
    \vspace{-10pt}
\end{figure}

 Since we cannot achieve 100\% correct translation in all cases, we examine a failure case, specifically the \texttt{Deformable Attention} in CUDA to BANG transcompilation, as shown in Figure~\ref{fig:error_translated}. This code snippet involves complex control flow, including multiple nested loops and conditional statements. As a result, neither the neural component (i.e., LLMs) of \name{} can handle such complex logical structures and conditional branches to the required SIMD intrinsic for BANG C, nor can the symbolic component (SMT solver) extract the mathematical constraints and perform code repairing. More powerful LLMs and advanced SMT analysis tools are promising solutions to this issue, which could further improve \name{}'s capability in handling such complex cases.

\section{Related Work}

\textbf{Rule-based approaches.}
These methods focus on transcompiling source programs to target languages using expert-defined rules~\cite{C2Rust, cxgo, HIPIFY, 10.1145/2400682.2400713}. For instance, C2Rust~\cite{C2Rust} and CxGo~\cite{cxgo} translate C code to Rust and Go, respectively, while HIPIFY~\cite{HIPIFY} converts CUDA to HIP. PPCG~\cite{10.1145/2400682.2400713} extracts the polyhedral model from the source program, applies predefined rules for scheduling and parallelization, and generates code from the transformed model.
These approaches are labor-intensive, relying on manual rules that require extensive knowledge of both the source and target languages, limiting their applicability to specific platforms. Moreover, they often struggle with irregular code structures, reducing their robustness and general applicability.

\textbf{Data-driven approaches.}
Data-driven methods, which train neural networks using supervised or unsupervised corpora, have gained prominence following the success of Neural Machine Translation (NMT). Tools like TransCoder~\cite{Roziere20NIPS} use sequence-to-sequence models for translating between C++, Java, and Python.
More recent works leverage LLMs (e.g., CodeX~\cite{chen2021evaluating}, StarCoder~\cite{li2023starcoder}, CodeGen~\cite{nijkamp2022codegen}, CodeT5~\cite{wang2021codet5}, CodeGeeX~\cite{zheng2023codegeex}, LLaMA~\cite{touvron2023llama}, Gemini~\cite{team2023gemini}, GPT-4~\cite{achiam2023gpt}, and OpenAI o1~\cite{openaio1}) for program translation, achieving superior performance.
However, while these methods reduce human effort, they often fail to guarantee the functional correctness of the translated code, especially across different DLS.

\textbf{Symbolic synthesis approaches.}
Symbolic synthesis approaches generate semantically equivalent code from input-output pairs or formal semantic specifications. For example, C2TACO~\cite{magalhaes2023c2taco} uses a guided enumerative synthesizer and automatically generated I/O examples to translate C tensor code into the TACO DSL~\cite{kjolstad2017taco}. Recent methods like MetaLift~\cite{bhatia_et_al:LIPIcs.ECOOP.2023.38} and Tenspiler~\cite{DBLP:conf/ecoop/QiuCBHSC24} adopt a more extensible approach by first converting the source program into a unified intermediate representation (IR) and then synthesizing target programs in the same IR domain.
These approaches reduce the need for manual transformation rules and are effective for small-scale projects. However, they rely on expensive search-based SMT solvers, which makes them less scalable for larger, real-world applications. Additionally, methods like MetaLift and Tenspiler require users to manually define the semantics of the target language using the specification IR, which is both challenging and error-prone.

In summary, existing approaches either demand significant manual effort, suffer from functional correctness issues, or have limited scalability, making them ineffective for automatically translating programs across DLS with different programming models.
In contrast, ~\name{} is the first work to tackle this challenge by proposing a novel neural-symbolic synthesis framework, where translation is performed through a series of LLM-assisted transformations, with functional equivalence ensured for each transformation via small-scale symbolic synthesis.

\section{Conclusion}
We propose a neural-symbolic synthesis approach, \name{}, for automatically translating tensor programs across heterogeneous deep learning systems with different programming models. \name{} conducts the program translation as a series of LLM-assisted transformation passes, where incorrect code snippets are repaired by small-scale symbolic synthesis. In addition to ensuring functional correctness, \name{} also employs hierarchical auto-tuning to improve the performance of translated programs. Experimental results on $4$ different DLS demonstrate that \name{} correctly translates different tensor programs at the accuracy of 95\% on average, and the performance of translated program achieves up to $2.0\times$ over vendor-provided manually-optimized libraries. Moreover, the programming productivity of DLS is improved by up to $96.0\times$.

\section*{Acknowledgments}

This work is partially supported by Strategic Priority Research Program of the Chinese Academy of Sciences (Grants No.XDB0660300, XDB0660301, XDB0660302), Science and Technology Major Special Program of Jiangsu (Grants No. BG2024028), the NSF of China (Grants No.U22A2028, 62302483, 6240073476, 62302482, 62302478), CAS Project for Young Scientists in Basic Research (YSBR-029) and Youth Innovation Promotion Association CAS.

\bibliographystyle{plain}
\bibliography{ref}

\end{document}